\pdfoutput=1

\documentclass[11pt]{article}

\usepackage[preprint]{acl}

\usepackage{times}
\usepackage{latexsym}

\usepackage[T1]{fontenc}

\usepackage[utf8]{inputenc}

\usepackage{microtype}

\usepackage{inconsolata}

\usepackage{graphicx}

\usepackage{subcaption} 
\usepackage{booktabs}

\title{Graph Linearization Methods for Reasoning on Graphs with Large Language Models}

\author{
 \textbf{Christos Xypolopoulos\textsuperscript{1,3}$^\dagger$},
 \textbf{Guokan Shang\textsuperscript{2}$^\dagger$},
  \textbf{Xiao Fei\textsuperscript{1}},\\
 \textbf{Giannis Nikolentzos\textsuperscript{4}},
 \textbf{Hadi Abdine\textsuperscript{2}},
 \textbf{Iakovos Evdaimon\textsuperscript{1}},\\
 \textbf{Michail Chatzianastasis\textsuperscript{1}},
 \textbf{Giorgos Stamou\textsuperscript{3}},
 \textbf{Michalis Vazirgiannis\textsuperscript{1,2}}
\\
\\
 \textsuperscript{1}Ecole Polytechnique,
 \textsuperscript{2}MBZUAI,
 \textsuperscript{3}NTUA,
 \textsuperscript{4}University of Peloponnese
\\
 \small{
   $^\dagger$Correspondence: \texttt{cxypolop@lix.polytechnique.fr, guokan.shang@mbzuai.ac.ae}
 }
}

\begin{document}
\maketitle

\begin{abstract}
Large language models have evolved to process multiple modalities beyond text, such as images and audio, which motivates us to explore how to effectively leverage them for graph reasoning tasks. 
The key question, therefore, is how to transform graphs into linear sequences of tokens---a process we term ``graph linearization''---so that LLMs can handle graphs naturally.
We consider that graphs should be linearized meaningfully to reflect certain properties of natural language text, such as local dependency and global alignment, in order to ease contemporary LLMs, trained on trillions of textual tokens, better understand graphs.
To achieve this, we developed several graph linearization methods based on graph centrality and degeneracy.
These methods are further enhanced using node relabeling techniques.
The experimental results demonstrate the effectiveness of our methods compared to the random linearization baseline.  
Our work introduces novel graph representations suitable for LLMs, contributing to the potential integration of graph machine learning with the trend of multimodal processing using a unified transformer model.	
\end{abstract}

\section{Introduction}

Transformer-based large pre-trained models have revolutionized machine learning research, demonstrating unprecedented performance across diverse data modalities and even a mixture of modalities, including image, audio, and text domains \citep{xu2023multimodal, yin2023survey}. 
In particular, large language models (LLMs) have shown promising results in arithmetic, symbolic, and logical reasoning tasks \citep{hendrycks2020measuring}.
Despite their success, the adaptation for processing graphs---an ubiquitous data structure that encapsulates rich structural and relational information---remains a comparably emerging and underdeveloped research direction, even if it has recently been gaining attention \citep{ye2023natural,fatemi2023talk,wang2024can}.
This asymmetry is due in large part to the inherent challenge of representing graphs as sequential tokens, in a manner conducive to the language modeling objectives typical of transformers, a challenge not encountered when dealing with the other modalities. 
This unique problem has encouraged us to investigate a key question: \textit{How can we represent graphs as linear sequences of tokens for transformers in a suitable way?} 
We refer to this research endeavor as \textit{Graph Linearization}.

Existing methods of using LLMs for graph machine learning tasks, such as graph reasoning and graph generation, represent entire graphs as either raw edge lists or natural language descriptions that adhere to adjacency matrices without any special treatment \citep{fatemi2023talk,wang2024can,yao2024exploring}.	
For example, a natural language description of the star graph $S$ might be: ``An undirected graph with nodes $a$, $b$, $c$, and $d$. Node $b$ is connected to $a$. Node $c$ is connected to $b$. Node $b$ is connected to $d$.'', its equivalent edge list representation is: ``[($b$, $a$), ($c$, $b$), ($b$, $d$)]''.
Either of the linearized representations is then appended with a task-specific question to form an LLM query prompt, e.g., ``Is there a cycle in this graph?''.
Other studies focus solely on node-level tasks \citep{zhao2023graphtext,ye2023natural}, centering the linearization around an ego-subgraph for a target node \citep{hamilton2017inductive}, where the neighboring graph structure and node features up to $k$-hop are described in the prompt.
However, there is a lack of studies investigating how to maintain the integrity of graph structures while efficiently transforming them into sequences suitable for LLMs.

\begin{figure}[ht]
\centering
\includegraphics[width=\columnwidth]{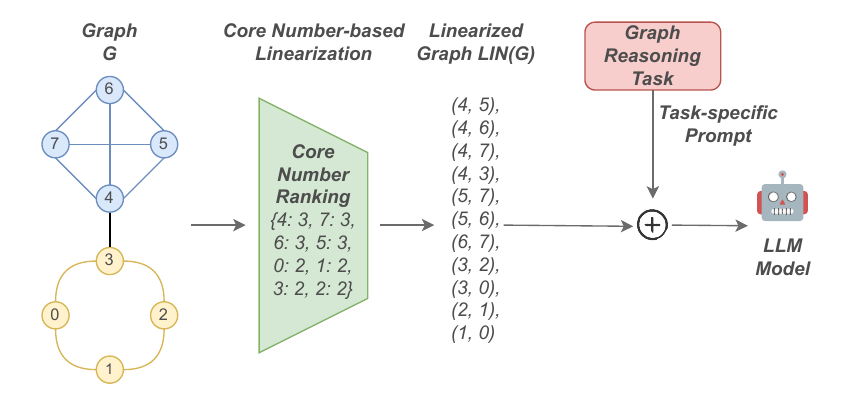}
\caption{An example of \textit{degree-based graph linearization} method. Given an input graph \textit{G}, we rank its nodes based on their degree and then explore the edges in that order. The resulting linearized graph is then combined with a task-specific prompt and passed into a LLM.}
\label{fig:architecture}
\end{figure}

Our research addresses this limitation. 
By relying on edge list representations as exemplified above, we study the performance impact on LLMs of various methods to order the edges in the list and rename interchangeable node labels, as shown in Figure \ref{fig:architecture}.	
We argue that if the linearization of graphs is conducted in a meaningful way, capturing properties similar to those found in natural language such as \textit{local dependency} and \textit{global alignment}, it will benefit contemporary LLMs by enhancing their ability to understand graphs, as they are trained on trillions of textual tokens.	
We define local dependency as the ability to predict the next (missing) token based on the previous (surrounding) context, as seen when a sequence of tokens naturally progresses in text, analogously to the distributional hypothesis of language \citep{joos1950description,harris1954distributional,firth1957synopsis}.
Global alignment involves starting sequences from tokens with similar characteristics, aligning the sequences to mimic how text samples begin or end with common words like ``The'' or ``In conclusion'' respectively.
Addressing this question helps identify LLM-suitable representations for graphs that potentially align with natural languages, unlocking new insights and applications in fields where graphs naturally represent data.		 
Graph linearization paves the way for extending transformer models' capabilities to graph data and unifying various graph tasks across domains, serving as a fundamental step toward building successful large graph models.
This also facilitates the integration of graph learning with the multi-modal processing trend and the development of cohesive AI systems using a unified transformer model.

In this work, we developed multiple general graph linearization methods that leverage graph centrality and degeneracy, enriched by node relabeling techniques to achieve the aforementioned natural language properties.
We conducted a series of inference experiments on various \textit{graph reasoning} tasks, which form the basis for a deeper understanding of graph structures.
Experimental results using Llama 3 models \citep{dubey2024llama} on synthetic datasets demonstrate the effectiveness of our methods compared to the random linearization baseline.
Our key findings are as follows:

    \noindent$\bullet$ \textbf{Edge ordering matters} – Structured graph linearization improves the graph reasoning capabilities of LLMs. \\
    \noindent$\bullet$ \textbf{Node labels contribute significantly} – Node labels based on graph features further improve performance. \\
    \noindent$\bullet$ \textbf{Task-specific linearizations} – Node-based and edge-based linearizations perform better in respective tasks.

\section{Related Work}

In this section, we first introduce previous efforts to use transformers for graph machine learning tasks.	
We then discuss the current trend of using modern LLMs for graph reasoning and generation.	
Finally, we explore linearization methods for graphs especially from specific domains.	

\subsection{Transformers for graph machine learning}

Transformers \citep{vaswani2017attention} have been successfully applied in various domains beyond text, demonstrating both versatility and effectiveness \citep{devlin2018bert}.
For instance, the Vision Transformer \citep{dosovitskiy2020image} has achieved remarkable performance in image classification tasks by treating images as sequences of patches, marking a shift from traditional CNN-based approaches.
Similarly, transformers have been used in speech recognition. Models like the Speech Transformer \citep{dong2018speech} apply self-attention mechanisms to process audio data as sequences, outperforming traditional RNN-based methods.

These successes have spurred interest in using modified transformers to replace the de facto GNN-based approaches for graph machine learning tasks.
Notably, Graphormer \citep{ying2021transformers} enables transformer to effectively capture the dependencies and relationships within a graph. It achieves this by integrating node centrality encoding and attention biases that account for the spatial distance between nodes. 
Graph Transformer (GT) \citep{dwivedi2020generalization} generalizes the transformer architecture for graph representation learning. GT introduces the concept of relative positional encodings to account for the pairwise distances between nodes in a graph. This approach allows the model to learn rich node representations that capture both local and global graph structures. 

In contrast to the above works, \citet{kim2022pure} show that by treating all nodes and edges as independent tokens and inputting them into a standard Transformer encoder without any graph-specific modifications, notable outcomes can be achieved both theoretically and practically.	
Results on molecular graphs for quantum chemical property prediction show that this approach outperforms all GNN baselines and achieves competitive performance compared to graph Transformer variants.	
Despite not applying any structural alterations to the Transformer, this approach still requires sophisticated token-wise node and edge embeddings to explicitly represent the connectivity structure.	

\subsection{LLMs for graph reasoning}
Following the recent success of LLMs in tasks beyond language processing \citep{hendrycks2020measuring}, several studies have explored the capacity of off-the-shelf LLMs for \textit{graph reasoning}.
While there is no clear consensus on the specific tasks, models are tested on understanding basic topological properties, such as graph size, node degree and connectivity, which form the foundation for a deeper understanding of graph structures \citep{zhang2023graph}.	
Using various prompting methods, these studies show that LLMs, even without fine-tuning, demonstrate preliminary graph reasoning abilities.	

Several studies have evaluated LLMs for graph reasoning at both the node and graph levels.
For example, NLGraph \citep{wang2024can} covers eight graph reasoning tasks of varying complexity, ranging from simple tasks like connectivity and shortest path to complex problems like maximum flow and simulating graph neural networks.	
This work also proposes two graph-specific prompting methods that achieve notable performance improvements.		
GraphQA \citep{fatemi2023talk} focuses on relatively simple tasks to measure the performance of pre-trained LLMs in edge existence, node degree, node count, edge count, connected nodes, and cycle checks.	
It shows that larger models generally perform better on graph reasoning, with graphs generated synthetically using various graph generators.	
Similar works include various studies that explore graph reasoning using different LLMs, prompting techniques, graph tasks, domains, and evaluation approaches \citep{chen2023exploring,guo2023gpt4graph,zhang2023llm4dyg,hu2023beyond,huang2024can,liu2023evaluating,das2023modality,yuan2024gracore,wu2024grapheval2000,skianis2024graph}.

Another line of research criticizes the above approach, arguing that solely using prompt engineering or in-context learning with frozen LLMs hinders achieving top performance in downstream graph tasks. Therefore, instruction-tuning or fine-tuning is necessary.	
The work of \citet{ye2023natural} preliminarily confirms this on the multi-class node classification task.
A prompt template is designed to describe both the neighbor graph structure and node features centered around a target node up to the 3-hop level. 
Similarly, \citet{zhao2023graphtext} draw inspiration from linguistic syntax trees. For a target node, the work converts its ego-subgraph into a graph syntax tree with branches describing the neighborhood’s ``label'' and ``feature'', which are then encoded as text in the prompt.
Results show that instruction-tuning performs much better than in-context learning, and is on par with GNN-based models.
Similar conclusions can be observed in recent studies on graph-level reasoning settings \citep{luo2024graphinstruct}.
Finally, \citet{perozzi2024let} introduce GraphToken, which trains an encoder to create continuous representations, rather than converting graphs into text tokens.

\subsection{LLMs for graph generation}

Graph generation involves creating graphs with specific properties, a process that holds significant real-world value in areas like drug discovery. This task is more challenging than graph reasoning.

To the best of our knowledge, \citet{yao2024exploring} was the first to show the preliminary abilities of LLMs in graph generation.	
The work experimented with graph generation in three settings: 1) Rule-based: generating graphs of basic structure types, given rules describing the desired structures, e.g., trees, cycles, etc.; 2) Distribution-based: generating graphs following a structural type distribution $p$, given a set of example graphs with the same distribution; 3) Property-based: generating molecule structures with specific properties, given example molecules in SMILES format.
Results show that LLMs have reasonably good abilities across the three tasks, and advanced prompting techniques do not necessarily lead to better performance.	
comment
Graph generation is critical for applications like drug discovery but presents greater challenges than graph reasoning. \citep{yao2024exploring} provide early insights into LLM-driven graph generation, evaluating rule-based, distribution-based, and property-based approaches. Their findings suggest that while LLMs can generate structurally meaningful graphs, advanced prompting does not always yield better performance, highlighting the need for more robust generation strategies.

\subsection{Linearization for specific graphs}
In deriving an ordering for graphs, topological sorting in graph theory examines the linear ordering of directed acyclic graphs, such that for every directed edge $(u, v)$, $u$ precedes $v$ in the ordering.
However, such graph traversal is node-centric, making edge information not encoded.

In other domains involving specific types of graphs, such as discourse graphs---a directed weakly connected graph reflecting discourse structure---nodes represent utterances, and edges represent discourse relations (e.g., elaboration, clarification, completion) within a conversation \cite{rennard2024leveraging}.	
The work of \citet{chernyavskiy-etal-2024-groundhog} proposes a linearization method for discourse graphs that arranges utterances chronologically, assigning unique identifiers to speakers, utterances, and addressees. It incorporates discourse relations and sentiment tokens to generate a structured sequence, using special tokens for clarity. This structured sequence is then used to train a BART \citep{lewis-etal-2020-bart} for dialogue generation.
Similarly, Abstract Meaning Representation (AMR) uses directed acyclic graphs to provide a structured semantic representation of language, incorporating semantic roles with annotated arguments and values where nodes represent concepts and edges represent semantic relations \citep{banarescu-etal-2013-abstract}. AMR corpora are usually linearized using the PENMAN-based notation \citep{patten-1993-book} as in the work of \citep{ribeiro-etal-2021-investigating} and \citep{hoyle-etal-2021-promoting} to fine-tune pre-trained language models to perform graph-to-text generation.
For citation networks, \citet{guo2023gpt4graph} have explored the Graph Modelling Language (GML) and Graph Markup Language (GraphML) for graph representation \citep{himsolt1997gml,brandes2013graph}. GML is a simple, human-readable format, while GraphML is XML-based and offers extensibility for complex applications.

\medskip

\noindent\textbf{The scope of our work}. Unlike the above linearization methods limited to specific types of graphs, where linearization can be naturally derived to some extent, we focus on general graphs.	
Furthermore, unlike previous works using LLMs for reasoning and generation tasks, where edge lists are directly leveraged without special treatment, we introduce various linearization methods for ordering the edges in the list and renaming interchangeable node labels to make them suitable for LLMs.
Although our work involves only graph reasoning experiments, our graph linearization methods are general and applicable to various scenarios.
This allows for the effective transformation of graph structures into sequences suitable for language models and has the potential to improve performance in both reasoning and generation tasks, with or without fine-tuning.

\section{Graph Linearization Methods} \label{sec:linearization}

This section describes the graph linearization approach, emphasizing the use of graph features to enhance graph reasoning with LLMs. 

Generally speaking, we define graph linearization as the process of representing graphs as linear sequences of tokens.	
In this work, we aim to identify the linearization approaches that will benefit LLMs by enhancing their ability to understand graphs.	
We argue that linearized graphs, represented as sequences of tokens, should capture properties similar to those in natural language, given the fact that LLMs are pre-trained on trillions of textual tokens.
Such properties should include local dependency and global alignment.

\smallskip
\noindent\textit{Local dependency} refers to the ability to predict the next (missing) token based on the previous (surrounding) context, within the token sequence of a single linearized graph.
This property is analogous to the fundamental \textit{distributional hypothesis} of language \citep{joos1950description,harris1954distributional,firth1957synopsis}, which states that words that occur in similar contexts tend to have similar meanings (or functions). 
This hypothesis suggests that given a new word, one should be able to figure out its meaning based on the contexts in which it is used. 
In fact, the masked and casual language modeling for training encoder-only \citep{devlin-etal-2019-bert} and decoder-only \citep{radford2019language} language models, can be seen as instantiations of this hypothesis.

\smallskip
\noindent\textit{Global alignment} refers to the alignment across the token sequences of different linearized graphs, ensuring that the corresponding tokens of both sequences match across their full length.
This property takes into account the overall structure of the sequences, reflecting the typical flow of text, where common words like ``The'' or ``In conclusion'' are used at the start or end of a sequence, guiding the alignment.
For example, linearization should always start from the node with the highest degree, and such nodes are all relabeled as index 0 for different graphs.

By relying on edge list, which defines a graph in terms of its individual connections, we study the performance impact on LLMs of various methods for ordering the edges in the list and renaming interchangeable node labels, as shown in Figure \ref{fig:architecture}.
More specifically, we leverage the advances in graph degeneracy and centrality as detailed in below to meet the local dependency property.

\medskip

\noindent \textbf{Graph Degeneracy} \citep{seidman1983network}.  
Let $G(V, E)$ be an undirected graph with  $n=\left\vert{V}\right\vert$ nodes and $m= \left\vert{E}\right\vert$ edges.
A $k$-core of $G$ is a maximal subgraph of $G$ in which every node $v$ has at least degree $k$.
The $k$-core decomposition of $G$ forms a hierarchy of nested subgraphs whose cohesiveness and size increase and decrease, respectively, with $k$.
Higher-level cores can be viewed as filtered versions of the graph that capture the most significant structural information.  
The \texttt{Core Number} of a node is the highest order of a core that contains the node.

\medskip

\noindent \textbf{Graph Centrality}.  
Graph centrality is a key concept in network analysis used to determine the influence or importance of nodes based on the structure of the graph. 
We consider two types of centrality, one based on the Degree of nodes and one based on PageRank.

\smallskip
\noindent\texttt{Degree} centrality is one of the simplest measures of node importance \citep{freeman1978centrality}.
It is defined as the number of edges incident to a node, making it a measure of local centrality that reflects the node's immediate connectivity within the graph.
For an undirected graph $G$ with $n=\left\vert{V}\right\vert$ nodes and $m= \left\vert{E}\right\vert$ edges, the degree centrality $D(v)$ of a vertex $v \in V$ is calculated as $D(v) = \sum_{u \in V} A_{vu}$ where $A$ is the adjacency matrix of the graph, and $A_{vu} = 1$ if there is an edge between vertices $v$ and $u$, and $0$ otherwise. 

\smallskip
\noindent\texttt{PageRank} is a more sophisticated centrality measure, originally developed by \citet{brin1998anatomy} for ranking web pages.
It extends the concept of degree centrality by considering not only the number of links a node has but also the importance of the nodes linking to it.
PageRank effectively captures the notion that connections from highly-ranked nodes contribute more to the ranking of a given node than connections from low-ranked nodes \citep{page1999pagerank}.
PageRank, although designed for directed graphs, can also be adapted for undirected ones by treating all edges as bidirectional.
The PageRank centrality $PR(v)$ of a node $v \in V$ is computed iteratively using the formula: $PR(v) = \frac{1 - \alpha}{|V|} + \alpha \sum_{u \in \mathcal{N}(v)} \frac{PR(u)}{\deg^+(u)}$ where $\alpha$ is a damping factor typically set to $0.85$, $\mathcal{N}(v)$ represents the set of nodes linking to $v$, and $\deg^+(u)$ is the out-degree of node $u$. 

\medskip

\noindent\textbf{Graph Linearization Implementation}. 
Our approach to capitalizing on the local dependency property involves the following steps.
Given a graph $G$, we initially rank the nodes by the centrality and degeneracy measures described previously.
Then, we begin exploring the nodes by descending order and list the edges connected to it, arranging them in a random order.
Each edge is represented as a node pair. 
In the case where two or more nodes share an equal value, the order is selected randomly.
After the ordering process has concluded, each edge list constitutes a sequence of tokens following a descending order of node importance. 

In addition to linearization methods, node relabeling is employed as a means to attempt the attainment of the global alignment property.
Specifically, node relabeling introduces an additional step to our procedure.
After ranking the nodes, their original labels are replaced with their respective positions in the ranking.
Consequently, the node with index 0 corresponds to the one with the highest core number, and so forth.
This approach may prove advantageous for the LLM by ensuring a consistent association between node indices and their respective importance properties.

Finally, we conducted experiments in which the edges were ordered directly, rather than the nodes.
This allows our linearization to directly capture relationships between edges, which can be essential for understanding complex graph structures.
To achieve this, each graph was transformed into its corresponding linegraph representation.
A linegraph $L(G)$ of a graph $G$ is the graph where each edge of $G$ is replaced by a node, and where two edges of $G$ are connected in $L(G)$ if they are incident in $G$.
Subsequently, the previously described processes were applied directly to  $L(G)$. 

\section{Experimental Setup}

In this section, we provide a detailed overview of the experimental setup, including the methodologies, resources, and evaluation frameworks used in our experiments.

\subsection{Datasets} 

\begin{figure*}[ht] 
\centering
\includegraphics[scale=0.48]{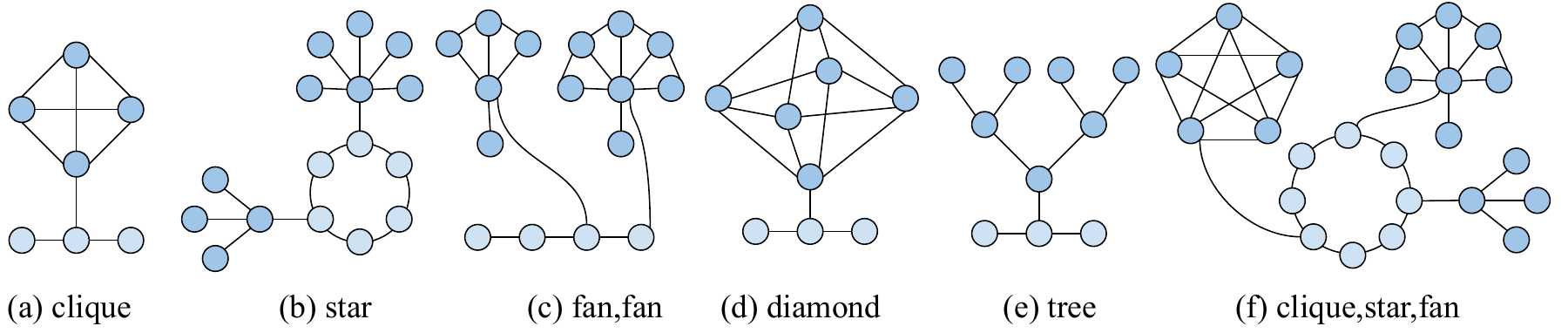}
\caption{Overview of GraphWave synthetic dataset.}
\label{fig:overview_graphwave}
\end{figure*}

To better explore the capabilities of LLMs on graph reasoning tasks, we utilized two synthetic datasets. 

\medskip
\noindent\textbf{GraphWave.}
First, we constructed a synthetic graph dataset using \texttt{GraphWave} \citep{donnat2018learning}.
This graph generator was originally developed for controlled experimentation and evaluation of node embedding techniques and measurement of structural equivalence on graphs with known network motifs. 
Motifs, in the context of network science, are sub-graphs that repeat themselves either within the same graph or across different graphs.
For our analysis, these structurally similar sub-graphs enable us to construct structure-related tasks, that allow us to assess the ability of language models to analyze and infer structural features within these graphs.

The generator operates by sequentially constructing a \textit{base} graph, that follows either a cycle or chain structure, and then attaching a number of \textit{motifs} that follow predetermined shapes---cliques, stars, fans, diamonds, and trees.
To cover a wider variety of structural complexities, we also include all combinations of two shapes, along with all combinations of three shapes with unique shapes per triplet. 
Example graphs can be found in Figure \ref{fig:overview_graphwave}.

For each combination of shapes, we generated 100 graphs, leading to a total of 3000 graphs.
This ensures sufficient variance in graph sizes and in the combinations of base and motif sub-graphs.	
The number of nodes in each graph shape is selected randomly, with the following constraints: base (3-21 nodes); clique, fan, and star (4-11 nodes); diamond (6 nodes); and tree (perfect binary trees with 3-6 levels).
The generated dataset contains an average of 32.33 nodes and 43.72 edges per graph.

\medskip
\noindent\textbf{GraphQA.}
Furthermore, we included \texttt{GraphQA} \citep{fatemi2023talk}, which is widely used to evaluate the graph reasoning capabilities of LLMs.
Similar to GraphWave, the dataset consists of randomly generated graphs derived from various graph generators, including Erdős–Rényi graphs \citep{erdos1959random}, scale-free networks (SFN) \citep{barabasi1999emergence}, the Barabási–Albert model (BA) \citep{albert2002statistical}, and the stochastic block model (SBM) \citep{holland1983stochastic}, along with star, path, and complete graph generators.
A total of 500 graphs were sampled for ER, BA, SFN, and SBM models, whereas 100 graphs were sampled for path, complete, and star graphs due to their lower structural variability.  
All generated graphs contained between 5 and 20 nodes.

\begin{table*}[ht]
  \centering
  \renewcommand{\arraystretch}{1}
  \resizebox{\textwidth}{!}{
\begin{tabular}{@{}lcccccccc||c@{}}
\toprule
& \textbf{Node Counting}  &  \textbf{Max Degree}  &  \textbf{Node Degree}  &  \textbf{Edge Existence}  &  \textbf{Diameter}  &  \textbf{Shortest Path}  &  \textbf{Path Existence} & \textbf{Motifs' Shape} & \textbf{Average} \\
\midrule
\textbf{\textit{Random Labeling}} &  &  &  &  &  &  &  & & \\
CoreNumber & 25.97 / 34.98  &  17.47 / 17.37  &  58.53 / \underline{56.79}  &  51.83 / \underline{55.29}  &  \underline{8.9} / 11.87  &  24.57 / \underline{17.84}  &  85.17 / 66.16 &  45.8 / 64.07 & 39.78 / 40.55\\
Degree & 28.0 / 36.98  &  \underline{27.63} / \underline{27.14}  &  \underline{60.83} / 52.22  &  47.6 / 53.72  &  7.97 / 10.17  &  \underline{27.8} / 15.81  &  84.63 / 69.12 &  \underline{\textbf{48.47}} / 64.8  & \underline{41.62} / 41.24\\
PageRank & 28.81 / \underline{39.18}  &  24.10 / 23.57  &  59.4 / 56.42  &  44.9 / 47.52  &  8.83 / 11.27  &  27.37 / 16.34  &  84.63 / \underline{72.69} &  44.93 / \underline{65.73}  & 40.37 / \underline{41.59}\\
LG\{CoreNumber\} & 21.9 / 27.38  &  19.00 / 16.47  &  46.23 / 42.68  &  52.47 / 50.12  &  8.73 / 12.0  &  23.03 / 16.21  &  83.87 / 64.22 &  44.4 / 63.13  & 37.45 / 36.53 \\
LG\{Degree\} & 20.4 / 26.51  &  \underline{27.63} / 18.87  &  48.07 / 47.92  &  \underline{55.47} / 52.65  &  8.7 / \underline{11.94}  &  26.3 / 17.64  &  84.8 / 62.29 &  44.33 / 61.17  & 39.46 / 37.37 \\
LG\{PageRank\} & \underline{29.1} / 27.28  &  18.83 / 17.97  &  46.8 / 40.61  &  47.9 / 49.02  &  8.47 / 12.14  &  24.73 / 16.04  &  \underline{87.07} / 70.52 &  42.4 / 60.0  & 38.16 / 36.7 \\
\midrule
\textbf{\textit{Node Relabeling}} &  &  &  &  &  &  &  & & \\
CoreNumber & 28.65 / 36.05  &  14.57 / 16.17 &  58.3 / \underline{\textbf{61.32}}  &  59.43 / 59.89  &  9.7 / 10.54  &  27.53 / \underline{\textbf{18.57}}  &  \underline{\textbf{88.0}} / \underline{\textbf{72.92}} &  44.3 / 64.57  & 41.31 / 42.5 \\
Degree & 31.44 / \underline{\textbf{43.85}}  & \underline{\textbf{29.40}} / \underline{\textbf{32.48}} &  62.93 / 56.79  &  58.5 / 55.09  &  \underline{\textbf{11.1}} / 11.24  &  30.3 / 16.31  &  82.0 / 72.79 &  \underline{48.37} / \underline{\textbf{76.3}}  & \textbf{\underline{44.26}} / \textbf{\underline{45.61}} \\
PageRank & \underline{\textbf{34.35}} / 38.01  &  26.07 / 25.24  &  \underline{\textbf{65.37}} / 56.92  &  50.83 / 46.08  &  10.47 / 11.17  &  \underline{\textbf{32.33}} / 17.67  &  82.13 / 72.89 &  47.6 / 68.63  & 43.64 / 42.08 \\
LG\{CoreNumber\} & 15.64 / 20.34  &  13.33 / 8.67 &  56.23 / 48.15  & \underline{\textbf{63.97}} / 55.55  &  8.5 / \underline{\textbf{12.5}}  &  27.6 / 17.67  &  85.83 / 70.66 &  45.0 / 50.9  & 39.51 / 35.56 \\
LG\{Degree\} & 23.59 / 31.94  &  19.83 / 16.11 &  48.9 / 47.85  &  60.03 / \underline{\textbf{63.05}}  &  10.1 / 12.14  &  29.63 / 17.31  &  84.8 / 70.36 &  41.07 / 54.23  & 39.74 / 39.12 \\
LG\{PageRank\} & 27.57 / 37.45  &  17.03 / 18.64 &  51.8 / 48.45  &  54.37 / 51.15  &  9.73 / 12.34  &  26.6 / 15.41  &  85.47 / 71.72 &  39.4 / 51.77  & 39.0 / 38.37\\
\midrule
\midrule
\textbf{\texttt{\Large Baseline}} & 32.46 / 34.6  &  15.13 / 12.94  &  39.36 / 45.99  &  36.99 / 53.25  &  7.99 / 12.38  &  22.73 / 16.57  &  86.98 / 68.05 &  43.53 / 66.8  & 34.86 / 34.86 \\
\bottomrule
\end{tabular}}
  \caption{Accuracy scores for all tasks on the \textbf{GraphWave} dataset using \textbf{Llama 3 8B}, including the overall average. Each task compares two node labeling schemes in zero-shot / one-shot prompting against the random \texttt{Baseline}. The linearization names represent the node ordering methods used (CoreNumber, Degree, or PageRank), while \textit{`LG\{*\}'} indicates graphs were transformed to the corresponding linegraph beforehand. Underlined scores denote the best-performing linearization method for each task, labeling, and X-shot combination; bold indicates task-wise best.}
 \label{table:graphwave_alltasks}

\end{table*}

\begin{table}[ht]
  \resizebox{\columnwidth}{!}{
\begin{tabular}{@{}lcccc@{}}
\toprule
& \textbf{Node Counting}  &  \textbf{Node Degree}  &  \textbf{Diameter}  &  \textbf{Motifs' Shape} \\
\midrule
\textbf{Random Labels} &  &  &  & \\
CoreNumber & 76.47 / 75.29  &  72.47 / \underline{63.75}  &  3.97 / 4.07  &  \textbf{\underline{59.1}} / \textbf{\underline{61.43}} \\
Degree & 78.2 / 80.96  &  73.93 / 63.69  &  2.43 / 4.57  &  58.07 / 61.4 \\
PageRank & 76.1 / 78.09  &  \underline{74.0} / 63.39  &  2.07 / 3.17  &  54.33 / 54.6 \\
LG\{CoreNumber\} & 77.8 / 77.36  &  68.83 / 57.02  &  4.77 / 6.97  &  46.8 / 56.23 \\
LG\{Degree\} & 78.9 / 81.16  &  72.23 / 62.65  &  4.57 / 7.5  &  51.5 / 54.2 \\
LG\{PageRank\} & \underline{84.83} / \underline{84.93}  &  72.4 / 61.25  &  \underline{4.93} / \underline{8.0}  &  44.7 / 43.93 \\
\midrule
\textbf{Node Relabeling} &  &  &   & \\
CoreNumber & \textbf{\underline{89.1}} / \textbf{\underline{86.06}}  &  75.87 / \textbf{\underline{70.99}}  &  8.47 / 11.3  &  54.97 / 53.4 \\
Degree & 84.83 / 82.53  &  76.0 / 69.39  &  7.27 / 11.8  &  \underline{58.23} / \underline{56.57} \\
PageRank & 84.77 / 81.13  &  73.93 / 67.59  &  8.63 / 12.84  &  58.1 / 47.4  \\
LG\{CoreNumber\} & 76.27 / 76.93  &  75.47 / 65.32  &  8.77 / 12.84  &  50.9 / 52.37 \\
LG\{Degree\} & 87.23 / 84.46  &  \textbf{\underline{76.63}} / 68.66  &  8.47 / \textbf{\underline{14.67}}  &  48.83 / 46.3 \\
LG\{PageRank\} & 86.23 / 84.86  &  76.43 / 63.32  &  \textbf{\underline{11.07}} / 13.7  &  47.2 / 41.1 \\
\midrule
\midrule
\textbf{Baseline} & 83.24 / 83.04  &  69.69 / 59.81  &  4.52 / 7.72  &  41.57 / 42.5 \\
\bottomrule
\end{tabular}}
\caption{Accuracy scores for a subset of tasks on the \textbf{GraphWave} dataset using \textbf{Llama 3 70B} as an ablation study. Notations remain the same as in Table \ref{table:graphwave_alltasks}.}
\label{table:graphwave_70b}
\end{table}

\subsection{Tasks}

Our experiments encompass a series of graph reasoning tasks, which can be broadly categorized into classification-based and numerical tasks.
Numerical tasks, ranging in difficulty, require the model to produce a numerical output, either through structural computation or counting-based inference.
This combination of tasks allows us to comprehensively evaluate LLMs' understanding of structural features and examine how different edge list orderings affect their performance. 

A fundamental task is \texttt{Node Counting}, where the LLM estimates the number of nodes.
In \texttt{Node Degree} calculation, the LLM determines the degree of a node.
A more advanced variant, \texttt{Maximum Degree} calculation, requires the LLM to internally calculate the degree of all nodes and then identify the maximum among them.

Beyond node-related tasks, we assess the ability to infer relational properties.
In \texttt{Edge Existence} and \texttt{Path Existence} tasks the LLM is given a randomly selected pair of nodes and must determine whether an edge or a connecting path exists between them, respectively.
In the \texttt{Shortest Path} task, the model must compute the length of the shortest path between two given nodes, requiring a deeper understanding of graph connectivity.
The \texttt{Diameter Estimation} task requires the model to determine the longest shortest path in the graph, showcasing global graph structure understanding. 

Finally, we evaluated \texttt{Motifs' Shape} classification, a dataset-specific task leveraging GraphWave's embedded structures, where the LLM is given definitions of the five motif types and asked is to identify which is present.

In every prompt, a node $v$ is represented by an incremental integer, while an edge between nodes $v$ and $u$ is denoted by the bracketed pair $(v, u)$. 
An edge list is expressed as a sequence of edges, sorted according to the scheme used in each linearization method. 
We tested both zero-shot and one-shot approaches, where a randomly selected graph from the dataset was used consistently as the one-shot example across all experiments. 
The prompt templates are provided in Appendix \ref{sec:prompts}.

\subsection{LLMs}

We used the 8B parameter Llama 3 Instruct \citep{dubey2024llama} with a temperature of $1\mathrm{e}{-3}$ and a sampling parameter of $1\mathrm{e}{-1}$ for more deterministic outputs to assess sensitivity to our linearization methods.
Experiments were conducted on an NVIDIA A5000.
Additionally, we conducted partial experiments with the 70B-parameter model to evaluate its impact on graph reasoning capabilities and to verify the consistency of our approach. 
Finally, to explore the impact of model family differences, we also include experiments over Qwen 2.5 14B-1M \cite{qwen2.5-1m} in the Appendix~\ref{sec:defaultlabeling}.

\subsection{Baselines}

For our comparisons, we consider only a random baseline. 
This baseline involves a fully random ordering of the edge list, where edges are arranged without following any inherent scheme.
To further eliminate structural biases, we also randomly shuffle the node labels.
This baseline is founded on the fact that we are working with general graphs, where default labels or ordering are neither predetermined nor necessarily provide meaningful information in real-world applications.
In addition, to mitigate the risk of skewed results, we applied five different random orderings and averaged their performance.
Our random ordering can be considered comparable to prior studies, which tend to preserve the inherent structure of the generator.

\begin{table*}[ht]
\renewcommand{\arraystretch}{1}
  \resizebox{\textwidth}{!}{
\begin{tabular}{@{}lccccccc||c@{}}
\toprule
& \textbf{Node Counting}  &  \textbf{Max Degree}  &  \textbf{Node Degree}  &  \textbf{Edge Existence}  &  \textbf{Diameter}  &  \textbf{Shortest Path}  &  \textbf{Path Existence} & \textbf{Average}\\
\midrule
\textbf{\textit{Random Labels}} &  &  &  &  &  & &\\
CoreNumber & 60.77 / 24.28  &  15.88 / 18.67  &  54.44 / 37.23  &  \underline{70.68} / 61.59  &  3.28 / \underline{\textbf{18.53}}  &  49.21 / \underline{\textbf{57.25}}  &  95.51 / 98.39 & 49.97 / 45.13 \\
Degree & \underline{62.18} / \underline{28.91}  &  25.6 / \underline{29.75}  &  54.99 / 29.4  &  68.02 / 58.6  &  \underline{3.39} / 14.82  &  \underline{\textbf{50.24}} / 53.34  &  94.39 / 98.71 & \underline{51.26} / 44.79 \\
PageRank & 62.12 / 28.31  &  \underline{26.03} / 28.94  &  \underline{56.23} / 28.05  &  68.39 / 59.26  &  3.16 / 15.25  &  29.34 / 47.3  &  94.51 / 98.68 & 48.54 / 43.68\\
LG\{CoreNumber\} & 61.14 / 23.94  &  10.81 / 25.89  &  49.09 / 36.65  &  69.26 / 58.83  &  2.9 / 16.97  &  42.08 / 42.35  &  \underline{95.92} / 98.71 & 47.31 / 43.33\\
LG\{Degree\} & 59.53 / 27.22  &  12.97 / 28.25  &  48.46 / 37.69  &  68.25 / 60.39  &  2.7 / 16.34  &  23.99 / 43.1  &  95.2 / \underline{\textbf{98.82}} & 44.44 / 44.54\\
LG\{PageRank\} & 60.66 / 26.24  &  11.59 / 28.65  &  47.51 / \underline{38.43}  &  69.49 / \underline{62.66}  &  2.67 / 17.43  &  40.78 / 45.6  &  94.31 / 98.79 & 46.72 / \underline{45.4} \\
\midrule
\textbf{\textit{Node Relabeling}} &  &  &  &  &  &  &  & \\
CoreNumber & 61.06 / 34.38  &  22.09 / 30.29  &  51.94 / 30.49  &  72.02 / 70.17  &  2.36 / 14.41  &  \underline{46.28} / \underline{52.85}  &  94.79 / 98.56 & 50.08 / 47.31 \\
Degree & 65.31 / \underline{\textbf{45.48}}  &  39.11 / \underline{\textbf{44.3}}  &  54.76 / 29.2  &  70.18 / \underline{\textbf{74.6}}  &  2.93 / 14.3  &  46.22 / 52.01  &  96.89 / 98.76 & \textbf{\underline{53.63}} / \textbf{\underline{51.24}}\\
PageRank & 64.71 / 44.97  &  \underline{\textbf{40.72}} / 43.15  &  \underline{\textbf{58.77}} / 28.65  &  69.97 / 70.83  &  \underline{\textbf{3.57}} / 14.13  &  26.17 / 49.54  &  \underline{\textbf{97.04}} / \underline{\textbf{98.82}} & 51.56 / 50.01 \\
LG\{CoreNumber\} & 63.56 / 32.54  &  22.69 / 33.49  &  47.48 / 29.14  &  \underline{\textbf{72.91}} / 59.15  &  1.93 / 15.1  &  24.33 / 43.84  &  96.52 / \underline{\textbf{98.82}} & 47.06 / 44.58 \\
LG\{Degree\} & 64.77 / 35.07  &  27.87 / 35.27  &  47.2 / \underline{\textbf{39.18}}  &  70.46 / 59.26  &  3.19 / 15.13  &  42.45 / 46.72  &  96.06 / 98.76 & 50.29 / 47.06 \\
LG\{PageRank\} & \underline{\textbf{69.37}} / 32.42  &  26.37 / 36.02  &  49.5 / 38.9  &  70.84 / 63.98  &  3.13 / \underline{15.19}  &  41.53 / 47.44  &  95.48 / 97.96 & 50.89 / 47.42 \\
\midrule
\midrule
\textbf{\texttt{\large Baseline}} & 66.28 / 27.38  &  9.28 / 20.37  &  49.78 / 35.86  &  67.02 / 58.99  &  2.94 / 14.98  &  37.7 / 44.55  &  95.83 / 98.51 & 46.98 / 42.95 \\
\bottomrule
\end{tabular}}
  \caption{Accuracy scores for all tasks on the \textbf{GraphQA} dataset using \textbf{Llama 3 8B}, , including the overall average. Notations remain the same as in Table \ref{table:graphwave_alltasks}.}
  \label{table:graphqa_allresults}
\end{table*}

\subsection{Evaluation}

We use exact accuracy to compare our methods, measuring the ratio of correct predictions as $\frac{1}{n} \sum_{i=1}^{n} I\left( y_i = \hat{y}_i \right)$, where $n$ is the number of graphs, $y_i$ the correct answer, and $\hat{y}_i$ the LLM's response.
For numerical tasks, we consider a result accurate only if it is an exact match. 
For the motifs' shape classification task, accuracy reflects the total across all shapes, requiring the predicted shape to appear at least once in the graph.

\section{Experimental Results Analysis}

In this section, we review the performance of our linearization methods across various tasks.

Tables \ref{table:graphwave_alltasks} and \ref{table:graphqa_allresults} present the performance of our methods on the GraphWave and GraphQA datasets with the Llama 3 8B model.
Additionally, to evaluate model scale, we report results for selected GraphWave tasks using the 70B model in Table \ref{table:graphwave_70b}.
The results are organized into three groups: one where node labels in the linearized graphs are randomly assigned, ensuring a fair comparison since, in real-world graphs, labels might be arbitrary; another where node labels are reindexed according to each method, as described in Section \ref{sec:linearization}; and finally, a comparison against the random baseline.
The performance related to the pseudo-random (default) node labels originally provided by synthetic graph generators is discussed in Appendix \ref{sec:defaultlabeling}.

Overall, across both datasets, our linearization methods consistently outperform the random baseline, highlighting the critical role of graph linearization, as evident in the average performance and across multiple individual tasks.
Notably, on the GraphQA dataset, the combination of degree-based ordering and node relabeling improves performance by approximately 35\% on the maximum node degree estimation task and by around 13\% on the shortest path task.
Similarly, on the GraphWave dataset, the combination of degree-based ordering and node relabeling enhances edge existence performance by roughly 26\%.
Comparing random and node relabeling reveals that ordering alone offers significant improvements over the baseline, while the additional information from structured relabeling further enhances accuracy on nearly all tasks.
Among all tasks, diameter estimation is the most challenging across both datasets, with consistently low performance, indicating LLMs struggle to infer global graph properties at this level of complexity.

Linegraph-based methods (LG\{*\})---where edges are reinterpreted as nodes---highlight the importance of edge-to-edge relationships.
While their overall average score is lower, they generally perform better in edge-based tasks, such as edge existence and path reasoning, by capturing interdependencies that might be less evident in traditional node-focused representations.
These findings suggest that a more suitable linearization approach may be necessary to fully exploit the benefits of the linegraph transformation.

Similarly, CoreNumber-based methods achieve better performance in edge-centric tasks, which can be attributed to its ability to capture the structural cohesiveness of a graph.
By emphasizing nodes embedded in densely connected subgraphs, core number ordering effectively preserves key connectivity patterns, making it particularly advantageous for reasoning about edge relationships.
In contrast, while Degree- and PageRank-based orderings demonstrate the most consistent performance across various tasks, their strengths are more pronounced in node-related tasks.

When moving from zero-shot to one-shot setting, we notice a performance loss on binary classification tasks like edge and path existence.
This decline may result from the model’s reliance on a single graph example, which does not fully capture the complexity and diversity of the dataset.
However, despite this drop, one-shot prompting remains effective for more complex tasks.

Finally, when comparing the results between the 8B and 70B versions of Llama, we observe a significant performance boost in the node counting and motif shape classification tasks with the larger model.
In contrast, the node degree task shows only a modest improvement.
Meanwhile, diameter estimation continues to exhibit very low accuracy, indicating that this task remains particularly challenging regardless of model size.


\section{Conclusion}

In this study, we investigated different graph linearization techniques for LLMs. The core challenge lies in converting graphs into linear token sequences while maintaining important structural features—like local dependencies and global coherence—akin to those in natural language, to help LLMs more effectively interpret graph-based data.
To this end, we developed and evaluated several linearization methods based on graph centrality and degeneracy.
Our experiments showed that graph linearization notably enhances the performance of LLMs on graph reasoning tasks, especially when paired with the node relabeling technique.
Our work presents novel graph representations tailored for LLMS, paving the way for integrating graph machine learning with the growing trend of multimodal processing through a unified transformer.

\section*{Limitations}
Our study has several limitations that future research could address.  
First, we considered only a limited set of structural features.
Key graph properties, such as community structures and connected components, were not incorporated, yet they could enhance the model’s ability to capture complex structural patterns.  
Second, the diversity of datasets is limited.
We conducted our experiments exclusively on synthetic graphs, and future work should explore a broader range of graph datasets to improve real-world generalization.  
Third, the range of LLMs evaluated was narrow.
We primarily focused on LLaMA 3, and future studies could investigate a wider variety of models, including those fine-tuned for coding or mathematical reasoning, to assess their impact on graph-based tasks. 
Fourth, our method inherits a fundamental limitation from the underlying LLMs: the input sequence is bounded by the model’s context window.
This imposes an upper limit on the number of edges that can be represented per graph.
In Appendix~\ref{sec:graph_length}, we analyze how this constraint translates to a maximum graph size under different edge densities and token budgets.
Lastly, our study does not investigate the impact of additional training to adapt models to our linearization methods, which could potentially enhance their graph reasoning capabilities.

\bibliography{main}

\begin{thebibliography}{59}
\providecommand{\natexlab}[1]{#1}

\bibitem[{Albert and Barab{\'a}si(2002)}]{albert2002statistical}
R{\'e}ka Albert and Albert-L{\'a}szl{\'o} Barab{\'a}si. 2002.
\newblock Statistical mechanics of complex networks.
\newblock \emph{Reviews of Modern Physics}, 74(1):47--97.

\bibitem[{Banarescu et~al.(2013)Banarescu, Bonial, Cai, Georgescu, Griffitt, Hermjakob, Knight, Koehn, Palmer, and Schneider}]{banarescu-etal-2013-abstract}
Laura Banarescu, Claire Bonial, Shu Cai, Madalina Georgescu, Kira Griffitt, Ulf Hermjakob, Kevin Knight, Philipp Koehn, Martha Palmer, and Nathan Schneider. 2013.
\newblock \href {https://aclanthology.org/W13-2322} {{A}bstract {M}eaning {R}epresentation for sembanking}.
\newblock In \emph{Proceedings of the 7th Linguistic Annotation Workshop and Interoperability with Discourse}, pages 178--186, Sofia, Bulgaria. Association for Computational Linguistics.

\bibitem[{Barabási and Albert(1999)}]{barabasi1999emergence}
Albert-László Barabási and R{\'e}ka Albert. 1999.
\newblock Emergence of scaling in random networks.
\newblock \emph{Science}, 286(5439):509--512.

\bibitem[{Brandes et~al.(2013)Brandes, Eiglsperger, Lerner, and Pich}]{brandes2013graph}
Ulrik Brandes, Markus Eiglsperger, Jürgen Lerner, and Christian Pich. 2013.
\newblock Graph markup language (graphml).
\newblock In Roberto Tamassia, editor, \emph{Handbook of graph drawing visualization}, Discrete mathematics and its applications, pages 517--541. CRC Press, Boca Raton [u.a.].

\bibitem[{Brin and Page(1998)}]{brin1998anatomy}
Sergey Brin and Lawrence Page. 1998.
\newblock The anatomy of a large-scale hypertextual web search engine.
\newblock In \emph{Computer Networks and ISDN Systems}, volume~30, pages 107--117. Elsevier.

\bibitem[{Chen et~al.(2023)Chen, Mao, Li, Jin, Wen, Wei, Wang, Yin, Fan, Liu et~al.}]{chen2023exploring}
Zhikai Chen, Haitao Mao, Hang Li, Wei Jin, Hongzhi Wen, Xiaochi Wei, Shuaiqiang Wang, Dawei Yin, Wenqi Fan, Hui Liu, et~al. 2023.
\newblock Exploring the potential of large language models (llms) in learning on graph.
\newblock In \emph{NeurIPS 2023 Workshop: New Frontiers in Graph Learning}.

\bibitem[{Chernyavskiy et~al.(2024)Chernyavskiy, Ostyakova, and Ilvovsky}]{chernyavskiy-etal-2024-groundhog}
Alexander Chernyavskiy, Lidiia Ostyakova, and Dmitry Ilvovsky. 2024.
\newblock \href {https://aclanthology.org/2024.codi-1.14} {{G}round{H}og: Dialogue generation using multi-grained linguistic input}.
\newblock In \emph{Proceedings of the 5th Workshop on Computational Approaches to Discourse (CODI 2024)}, pages 149--160, St. Julians, Malta. Association for Computational Linguistics.

\bibitem[{Chiang et~al.(2019)Chiang, Liu, Si, Li, Bengio, and Hsieh}]{chiang2019cluster}
Wei-Lin Chiang, Xuanqing Liu, Si~Si, Yang Li, Samy Bengio, and Cho-Jui Hsieh. 2019.
\newblock \href {https://arxiv.org/abs/1907.04931} {Cluster-gcn: An efficient algorithm for training deep and large graph convolutional networks}.
\newblock \emph{arXiv preprint arXiv:1907.04931}.

\bibitem[{Das et~al.(2023)Das, Gupta, Srivastava, and Kang}]{das2023modality}
Debarati Das, Ishaan Gupta, Jaideep Srivastava, and Dongyeop Kang. 2023.
\newblock Which modality should i use--text, motif, or image?: Understanding graphs with large language models.
\newblock \emph{arXiv preprint arXiv:2311.09862}.

\bibitem[{Devlin et~al.(2018)Devlin, Chang, Lee, and Toutanova}]{devlin2018bert}
Jacob Devlin, Ming-Wei Chang, Kenton Lee, and Kristina Toutanova. 2018.
\newblock Bert: Pre-training of deep bidirectional transformers for language understanding.
\newblock \emph{arXiv preprint arXiv:1810.04805}.

\bibitem[{Devlin et~al.(2019)Devlin, Chang, Lee, and Toutanova}]{devlin-etal-2019-bert}
Jacob Devlin, Ming-Wei Chang, Kenton Lee, and Kristina Toutanova. 2019.
\newblock \href {https://doi.org/10.18653/v1/N19-1423} {{BERT}: Pre-training of deep bidirectional transformers for language understanding}.
\newblock In \emph{Proceedings of the 2019 Conference of the North {A}merican Chapter of the Association for Computational Linguistics: Human Language Technologies, Volume 1 (Long and Short Papers)}, pages 4171--4186, Minneapolis, Minnesota. Association for Computational Linguistics.

\bibitem[{Dong et~al.(2018)Dong, Xu, and Xu}]{dong2018speech}
Linhao Dong, Shuang Xu, and Bo~Xu. 2018.
\newblock Speech-transformer: a no-recurrence sequence-to-sequence model for speech recognition.
\newblock In \emph{2018 IEEE international conference on acoustics, speech and signal processing (ICASSP)}, pages 5884--5888. IEEE.

\bibitem[{Donnat et~al.(2018)Donnat, Zitnik, Hallac, and Leskovec}]{donnat2018learning}
Claire Donnat, Marinka Zitnik, David Hallac, and Jure Leskovec. 2018.
\newblock Learning structural node embeddings via diffusion wavelets.
\newblock In \emph{Proceedings of the 24th ACM SIGKDD international conference on knowledge discovery \& data mining}, pages 1320--1329.

\bibitem[{Dosovitskiy et~al.(2020)Dosovitskiy, Beyer, Kolesnikov, Weissenborn, Zhai, Unterthiner, Dehghani, Minderer, Heigold, Gelly et~al.}]{dosovitskiy2020image}
Alexey Dosovitskiy, Lucas Beyer, Alexander Kolesnikov, Dirk Weissenborn, Xiaohua Zhai, Thomas Unterthiner, Mostafa Dehghani, Matthias Minderer, Georg Heigold, Sylvain Gelly, et~al. 2020.
\newblock An image is worth 16x16 words: Transformers for image recognition at scale.
\newblock \emph{arXiv preprint arXiv:2010.11929}.

\bibitem[{Dubey et~al.(2024)Dubey, Jauhri, Pandey, Kadian, Al-Dahle, Letman, Mathur, Schelten, Yang, Fan et~al.}]{dubey2024llama}
Abhimanyu Dubey, Abhinav Jauhri, Abhinav Pandey, Abhishek Kadian, Ahmad Al-Dahle, Aiesha Letman, Akhil Mathur, Alan Schelten, Amy Yang, Angela Fan, et~al. 2024.
\newblock The llama 3 herd of models.
\newblock \emph{arXiv preprint arXiv:2407.21783}.

\bibitem[{Dwivedi and Bresson(2020)}]{dwivedi2020generalization}
Vijay~Prakash Dwivedi and Xavier Bresson. 2020.
\newblock A generalization of transformer networks to graphs.
\newblock \emph{arXiv preprint arXiv:2012.09699}.

\bibitem[{Erdős and Rényi(1959)}]{erdos1959random}
Paul Erdős and Alfréd Rényi. 1959.
\newblock On random graphs.
\newblock \emph{Publicationes Mathematicae}, 6:290--297.

\bibitem[{Fatemi et~al.(2023)Fatemi, Halcrow, and Perozzi}]{fatemi2023talk}
Bahare Fatemi, Jonathan Halcrow, and Bryan Perozzi. 2023.
\newblock Talk like a graph: Encoding graphs for large language models.
\newblock \emph{arXiv preprint arXiv:2310.04560}.

\bibitem[{Firth(1957)}]{firth1957synopsis}
John~R Firth. 1957.
\newblock A synopsis of linguistic theory, 1930-1955.
\newblock \emph{Studies in linguistic analysis}.

\bibitem[{Freeman(1978)}]{freeman1978centrality}
Linton~C Freeman. 1978.
\newblock Centrality in social networks conceptual clarification.
\newblock \emph{Social networks}, 1(3):215--239.

\bibitem[{Guo et~al.(2023)Guo, Du, Liu, Zhou, He, and Han}]{guo2023gpt4graph}
Jiayan Guo, Lun Du, Hengyu Liu, Mengyu Zhou, Xinyi He, and Shi Han. 2023.
\newblock Gpt4graph: Can large language models understand graph structured data? an empirical evaluation and benchmarking.
\newblock \emph{arXiv preprint arXiv:2305.15066}.

\bibitem[{Hamilton et~al.(2017)Hamilton, Ying, and Leskovec}]{hamilton2017inductive}
Will Hamilton, Zhitao Ying, and Jure Leskovec. 2017.
\newblock Inductive representation learning on large graphs.
\newblock \emph{Advances in neural information processing systems}, 30.

\bibitem[{Harris(1954)}]{harris1954distributional}
Zellig~S Harris. 1954.
\newblock Distributional structure.
\newblock \emph{Word}, 10(2-3):146--162.

\bibitem[{Hendrycks et~al.(2020)Hendrycks, Burns, Basart, Zou, Mazeika, Song, and Steinhardt}]{hendrycks2020measuring}
Dan Hendrycks, Collin Burns, Steven Basart, Andy Zou, Mantas Mazeika, Dawn Song, and Jacob Steinhardt. 2020.
\newblock Measuring massive multitask language understanding.
\newblock \emph{arXiv preprint arXiv:2009.03300}.

\bibitem[{Himsolt(1997)}]{himsolt1997gml}
Michael Himsolt. 1997.
\newblock Gml: Graph modelling language.
\newblock \emph{University of Passau}.

\bibitem[{Holland et~al.(1983)Holland, Laskey, and Leinhardt}]{holland1983stochastic}
Paul~W. Holland, Kathryn~B. Laskey, and Samuel Leinhardt. 1983.
\newblock Stochastic blockmodels: First steps.
\newblock \emph{Social Networks}, 5(2):109--137.

\bibitem[{Hoyle et~al.(2021)Hoyle, Marasovi{\'c}, and Smith}]{hoyle-etal-2021-promoting}
Alexander~Miserlis Hoyle, Ana Marasovi{\'c}, and Noah~A. Smith. 2021.
\newblock \href {https://doi.org/10.18653/v1/2021.findings-acl.82} {Promoting graph awareness in linearized graph-to-text generation}.
\newblock In \emph{Findings of the Association for Computational Linguistics: ACL-IJCNLP 2021}, pages 944--956, Online. Association for Computational Linguistics.

\bibitem[{Hu et~al.(2020)Hu, Fey, Zitnik, Dong, Ren, Liu, Catasta, and Leskovec}]{hu2020open}
Weihua Hu, Matthias Fey, Marinka Zitnik, Yuxiao Dong, Hongyu Ren, Bowen Liu, Michele Catasta, and Jure Leskovec. 2020.
\newblock \href {https://arxiv.org/abs/2005.00687} {Open graph benchmark: Datasets for machine learning on graphs}.
\newblock In \emph{Advances in Neural Information Processing Systems}, volume~33, pages 22118--22133.

\bibitem[{Hu et~al.(2023)Hu, Zhang, and Zhao}]{hu2023beyond}
Yuntong Hu, Zheng Zhang, and Liang Zhao. 2023.
\newblock Beyond text: A deep dive into large language models' ability on understanding graph data.
\newblock \emph{arXiv preprint arXiv:2310.04944}.

\bibitem[{Huang et~al.(2024)Huang, Zhang, Mei, and Ma}]{huang2024can}
Jin Huang, Xingjian Zhang, Qiaozhu Mei, and Jiaqi Ma. 2024.
\newblock Can llms effectively leverage graph structural information through prompts, and why?
\newblock \emph{Transactions on Machine Learning Research}.

\bibitem[{Joos(1950)}]{joos1950description}
Martin Joos. 1950.
\newblock Description of language design.
\newblock \emph{The Journal of the Acoustical Society of America}, 22(6):701--707.

\bibitem[{Kim et~al.(2022)Kim, Nguyen, Min, Cho, Lee, Lee, and Hong}]{kim2022pure}
Jinwoo Kim, Dat Nguyen, Seonwoo Min, Sungjun Cho, Moontae Lee, Honglak Lee, and Seunghoon Hong. 2022.
\newblock Pure transformers are powerful graph learners.
\newblock \emph{Advances in Neural Information Processing Systems}, 35:14582--14595.

\bibitem[{Kipf and Welling(2016)}]{kipf2016semi}
Thomas~N Kipf and Max Welling. 2016.
\newblock \href {https://arxiv.org/abs/1603.08861} {Semi-supervised classification with graph convolutional networks}.
\newblock \emph{arXiv preprint arXiv:1603.08861}.

\bibitem[{Lewis et~al.(2020)Lewis, Liu, Goyal, Ghazvininejad, Mohamed, Levy, Stoyanov, and Zettlemoyer}]{lewis-etal-2020-bart}
Mike Lewis, Yinhan Liu, Naman Goyal, Marjan Ghazvininejad, Abdelrahman Mohamed, Omer Levy, Veselin Stoyanov, and Luke Zettlemoyer. 2020.
\newblock \href {https://doi.org/10.18653/v1/2020.acl-main.703} {{BART}: Denoising sequence-to-sequence pre-training for natural language generation, translation, and comprehension}.
\newblock In \emph{Proceedings of the 58th Annual Meeting of the Association for Computational Linguistics}, pages 7871--7880, Online. Association for Computational Linguistics.

\bibitem[{Liu and Wu(2023)}]{liu2023evaluating}
Chang Liu and Bo~Wu. 2023.
\newblock Evaluating large language models on graphs: Performance insights and comparative analysis.
\newblock \emph{arXiv preprint arXiv:2308.11224}.

\bibitem[{Luo et~al.(2024)Luo, Song, Huang, Lian, Zhang, Jiang, Xie, and Jin}]{luo2024graphinstruct}
Zihan Luo, Xiran Song, Hong Huang, Jianxun Lian, Chenhao Zhang, Jinqi Jiang, Xing Xie, and Hai Jin. 2024.
\newblock Graphinstruct: Empowering large language models with graph understanding and reasoning capability.
\newblock \emph{arXiv preprint arXiv:2403.04483}.

\bibitem[{Page et~al.(1999)Page, Brin, Motwani, and Winograd}]{page1999pagerank}
Lawrence Page, Sergey Brin, Rajeev Motwani, and Terry Winograd. 1999.
\newblock The pagerank citation ranking: Bringing order to the web.
\newblock Technical report, Stanford InfoLab.

\bibitem[{Patten(1993)}]{patten-1993-book}
Terry Patten. 1993.
\newblock \href {https://aclanthology.org/J93-1011} {Book reviews: Text generation and systemic-functional linguistics: Experiences from {E}nglish and {J}apanese}.
\newblock \emph{Computational Linguistics}, 19(1).

\bibitem[{Perozzi et~al.(2024)Perozzi, Fatemi, Zelle, Tsitsulin, Kazemi, Al-Rfou, and Halcrow}]{perozzi2024let}
Bryan Perozzi, Bahare Fatemi, Dustin Zelle, Anton Tsitsulin, Mehran Kazemi, Rami Al-Rfou, and Jonathan Halcrow. 2024.
\newblock Let your graph do the talking: Encoding structured data for llms.
\newblock \emph{arXiv preprint arXiv:2402.05862}.

\bibitem[{Radford et~al.(2019)Radford, Wu, Child, Luan, Amodei, Sutskever et~al.}]{radford2019language}
Alec Radford, Jeffrey Wu, Rewon Child, David Luan, Dario Amodei, Ilya Sutskever, et~al. 2019.
\newblock Language models are unsupervised multitask learners.
\newblock \emph{OpenAI blog}, 1(8):9.

\bibitem[{Rennard et~al.(2024)Rennard, Shang, Vazirgiannis, and Hunter}]{rennard2024leveraging}
Virgile Rennard, Guokan Shang, Michalis Vazirgiannis, and Julie Hunter. 2024.
\newblock \href {https://arxiv.org/abs/2405.11055} {Leveraging discourse structure for extractive meeting summarization}.
\newblock \emph{Preprint}, arXiv:2405.11055.

\bibitem[{Ribeiro et~al.(2021)Ribeiro, Schmitt, Sch{\"u}tze, and Gurevych}]{ribeiro-etal-2021-investigating}
Leonardo F.~R. Ribeiro, Martin Schmitt, Hinrich Sch{\"u}tze, and Iryna Gurevych. 2021.
\newblock \href {https://doi.org/10.18653/v1/2021.nlp4convai-1.20} {Investigating pretrained language models for graph-to-text generation}.
\newblock In \emph{Proceedings of the 3rd Workshop on Natural Language Processing for Conversational AI}, pages 211--227, Online. Association for Computational Linguistics.

\bibitem[{Seidman(1983)}]{seidman1983network}
Stephen~B Seidman. 1983.
\newblock \href {https://doi.org/10.1016/0378-8733(83)90028-X} {Network structure and minimum degree}.
\newblock \emph{Social networks}, 5(3):269--287.

\bibitem[{Skianis et~al.(2024)Skianis, Nikolentzos, and Vazirgiannis}]{skianis2024graph}
Konstantinos Skianis, Giannis Nikolentzos, and Michalis Vazirgiannis. 2024.
\newblock Graph reasoning with large language models via pseudo-code prompting.
\newblock \emph{arXiv preprint arXiv:2409.17906}.

\bibitem[{Vaswani et~al.(2017)Vaswani, Shazeer, Parmar, Uszkoreit, Jones, Gomez, Kaiser, and Polosukhin}]{vaswani2017attention}
Ashish Vaswani, Noam Shazeer, Niki Parmar, Jakob Uszkoreit, Llion Jones, Aidan~N Gomez, {\L}ukasz Kaiser, and Illia Polosukhin. 2017.
\newblock Attention is all you need.
\newblock \emph{Advances in neural information processing systems}, 30.

\bibitem[{Wang et~al.(2024)Wang, Feng, He, Tan, Han, and Tsvetkov}]{wang2024can}
Heng Wang, Shangbin Feng, Tianxing He, Zhaoxuan Tan, Xiaochuang Han, and Yulia Tsvetkov. 2024.
\newblock Can language models solve graph problems in natural language?
\newblock \emph{Advances in Neural Information Processing Systems}, 36.

\bibitem[{Wu et~al.(2024)Wu, Chen, Corcoran, Sra, and Singh}]{wu2024grapheval2000}
Qiming Wu, Zichen Chen, Will Corcoran, Misha Sra, and Ambuj~K Singh. 2024.
\newblock Grapheval2000: Benchmarking and improving large language models on graph datasets.
\newblock \emph{arXiv preprint arXiv:2406.16176}.

\bibitem[{Xu et~al.(2023)Xu, Zhu, and Clifton}]{xu2023multimodal}
Peng Xu, Xiatian Zhu, and David~A Clifton. 2023.
\newblock Multimodal learning with transformers: A survey.
\newblock \emph{IEEE Transactions on Pattern Analysis and Machine Intelligence}.

\bibitem[{Yang et~al.(2025)Yang, Yu, Li, Liu, Huang, Huang, Jiang, Tu, Zhang, Zhou et~al.}]{qwen2.5-1m}
An~Yang, Bowen Yu, Chengyuan Li, Dayiheng Liu, Fei Huang, Haoyan Huang, Jiandong Jiang, Jianhong Tu, Jianwei Zhang, Jingren Zhou, et~al. 2025.
\newblock Qwen2.5-1m technical report.
\newblock \emph{arXiv preprint arXiv:2501.15383}.

\bibitem[{Yang et~al.(2018)Yang, Cohen, and Salakhutdinov}]{yang2018benchmarking}
Zhilin Yang, William~W Cohen, and Ruslan Salakhutdinov. 2018.
\newblock \href {https://arxiv.org/abs/1811.05868} {Benchmarking graph neural networks}.
\newblock \emph{arXiv preprint arXiv:1811.05868}.

\bibitem[{Yao et~al.(2024)Yao, Wang, Zhang, Qin, Zhang, Chu, Yang, Zhu, and Mei}]{yao2024exploring}
Yang Yao, Xin Wang, Zeyang Zhang, Yijian Qin, Ziwei Zhang, Xu~Chu, Yuekui Yang, Wenwu Zhu, and Hong Mei. 2024.
\newblock Exploring the potential of large language models in graph generation.
\newblock \emph{arXiv preprint arXiv:2403.14358}.

\bibitem[{Ye et~al.(2023)Ye, Zhang, Wang, Xu, and Zhang}]{ye2023natural}
Ruosong Ye, Caiqi Zhang, Runhui Wang, Shuyuan Xu, and Yongfeng Zhang. 2023.
\newblock Natural language is all a graph needs.
\newblock \emph{arXiv preprint arXiv:2308.07134}.

\bibitem[{Yin et~al.(2023)Yin, Fu, Zhao, Li, Sun, Xu, and Chen}]{yin2023survey}
Shukang Yin, Chaoyou Fu, Sirui Zhao, Ke~Li, Xing Sun, Tong Xu, and Enhong Chen. 2023.
\newblock A survey on multimodal large language models.
\newblock \emph{arXiv preprint arXiv:2306.13549}.

\bibitem[{Ying et~al.(2021)Ying, Cai, Luo, Zheng, Ke, He, Shen, and Liu}]{ying2021transformers}
Chengxuan Ying, Tianle Cai, Shengjie Luo, Shuxin Zheng, Guolin Ke, Di~He, Yanming Shen, and Tie-Yan Liu. 2021.
\newblock Do transformers really perform badly for graph representation?
\newblock \emph{Advances in Neural Information Processing Systems}, 34:28877--28888.

\bibitem[{Yuan et~al.(2024)Yuan, Liu, Wang, and Qin}]{yuan2024gracore}
Zike Yuan, Ming Liu, Hui Wang, and Bing Qin. 2024.
\newblock Gracore: Benchmarking graph comprehension and complex reasoning in large language models.
\newblock \emph{arXiv preprint arXiv:2407.02936}.

\bibitem[{Zhang et~al.(2023{\natexlab{a}})Zhang, Wang, Zhang, Li, Qin, Wu, and Zhu}]{zhang2023llm4dyg}
Zeyang Zhang, Xin Wang, Ziwei Zhang, Haoyang Li, Yijian Qin, Simin Wu, and Wenwu Zhu. 2023{\natexlab{a}}.
\newblock Llm4dyg: Can large language models solve problems on dynamic graphs?
\newblock \emph{arXiv preprint arXiv:2310.17110}.

\bibitem[{Zhang et~al.(2023{\natexlab{b}})Zhang, Li, Zhang, Qin, Wang, and Zhu}]{zhang2023graph}
Ziwei Zhang, Haoyang Li, Zeyang Zhang, Yijian Qin, Xin Wang, and Wenwu Zhu. 2023{\natexlab{b}}.
\newblock Graph meets llms: Towards large graph models.
\newblock In \emph{NeurIPS 2023 Workshop: New Frontiers in Graph Learning}.

\bibitem[{Zhao et~al.(2023)Zhao, Zhuo, Shen, Qu, Liu, Bronstein, Zhu, and Tang}]{zhao2023graphtext}
Jianan Zhao, Le~Zhuo, Yikang Shen, Meng Qu, Kai Liu, Michael Bronstein, Zhaocheng Zhu, and Jian Tang. 2023.
\newblock Graphtext: Graph reasoning in text space.
\newblock \emph{arXiv preprint arXiv:2310.01089}.

\bibitem[{Zitnik and Leskovec(2017)}]{zitnik2017predicting}
Marinka Zitnik and Jure Leskovec. 2017.
\newblock \href {https://arxiv.org/abs/1707.04638} {Predicting multicellular function through multi-layer tissue networks}.
\newblock \emph{arXiv preprint arXiv:1707.04638}.

\end{thebibliography}

\appendix

\section{Additional Experiments}\label{sec:defaultlabeling}

\noindent \textbf{Pseudo-random linearization.}
We also investigated the performance of utilizing the edge ordering directly provided by the graph generator.
Most non-random graph generators create graphs procedurally, inherently embedding structural information within the edge list order.
For instance, in GraphWave, the process begins with a base graph and subsequently attaches motifs, making it easier to distinguish between different structures.
We hypothesize that this embedded structural knowledge will enhance task performance and boost the LLM's capabilities.
However, generating such structure-aware edge lists requires an understanding of the graph construction process, which may not be feasible for real-world applications involving larger and more complex graphs.

The results of both datasets are presented in Table \ref{tab:defaultlabels}.
For the GraphWave dataset, the default edge ordering shows mixed performance compared to the random ordering baseline.
When combined with structured edge ordering (Table \ref{table:graphwave_alltasks}), accuracy improves consistently across tasks, except for path existence.
When comparing default labeling with structured node labeling, performance generally improves, though default labeling remains competitive in certain tasks.
For the GraphQA dataset, the default edge ordering performs significantly better than the random ordering across all tasks except path existence.
In this case, the default ordering proves to be particularly robust, making it challenging for structured edge ordering to achieve higher accuracy.
Even compared to structured edge ordering (Table \ref{table:graphqa_allresults}), default ordering often maintains a performance advantage, highlighting its effectiveness in this dataset.

\begin{table*}[ht]
  \centering
  \resizebox{\textwidth}{!}{
\begin{tabular}{@{}lcccccccc@{}}
\toprule
& \textbf{Node Counting}  &  \textbf{Max Degree}  &  \textbf{Node Degree}  &  \textbf{Edge Existence}  &  \textbf{Diameter}  &  \textbf{Shortest Path}  &  \textbf{Path Existence}  & \textbf{Motifs' Shape} \\
\toprule
\textbf{Default Labels} &  &  &  &  &  &  &  & \\
CoreNumber & 25.06 / 37.41  &  17.87 / 16.67  &  60.9 / \textbf{62.45}  &  54.43 / \textbf{57.69}  &  7.93 / 11.74  &  29.97 / \textbf{20.74}  &  83.1 / 70.32 &  43.43 / 63.23 \\
Degree & 32.66 / \textbf{51.18}  &  \textbf{28.37} / 26.24  &  60.77 / 57.79  &  55.5 / 49.22  &  8.97 / 11.54  &  \textbf{31.7} / 15.97  &  78.9 / 67.36 &  47.8 / 61.4 \\
PageRank & 35.77 / 49.38  &  24.40 / \textbf{26.88} &  \textbf{63.43} / 56.42  &  49.4 / 53.08  &  8.63 / 11.5  &  31.33 / 18.24  &  81.3 / 73.99 &  48.03 / \textbf{65.27} \\
LG\{CoreNumber\} & 19.79 / 22.21  &  21.60 / 13.40  &  55.37 / 47.75  &  \textbf{59.4} / 53.35  &  10.07 / 11.9  &  27.03 / 19.74  &  79.4 / 73.06 &  44.43 / 56.23 \\
LG\{Degree\} & 24.6 / 32.21  &  23.23 / 17.84  &  49.33 / 41.35  &  56.4 / 52.72  &  \textbf{10.67} / 11.04  &  29.33 / 19.47  &  79.67 / 71.76 &  43.87 / 52.73 \\
LG\{PageRank\} & 34.83 / 41.78  &  17.90 / 19.14  &  54.3 / 50.18  &  48.9 / 52.15  &  8.77 / \textbf{12.17}  &  27.63 / 16.47  &  79.77 / 76.59 &  43.07 / 48.63 \\
\midrule
Default Ordering & \textbf{36.75} / 44.38  &  9.77 / 10.17  &  47.83 / 41.18  &  37.23 / 47.48  &  7.3 / 11.5  &  30.3 / 16.67  &  \textbf{85.5} / \textbf{82.79} &  \textbf{54.63} / 55.73 \\
\bottomrule
\end{tabular}}

\medskip

\resizebox{\textwidth}{!}{
\begin{tabular}{@{}lccccccccc@{}}
\toprule
& \textbf{Node Counting}  &  \textbf{Max Degree}  &  \textbf{Node Degree}  &  \textbf{Edge Existence}  &  \textbf{Diameter}  &  \textbf{Shortest Path}  &  \textbf{Path Existence} \\
\toprule
\textbf{Default Labels} &  &  &  &  &  &  &  & & \\
CoreNumber & 61.23 / 38.52  &  22.4 / 34.35  &  52.72 / \textbf{44.51}  &  72.19 / 68.07  &  1.58 / 9.29  &  \textbf{46.65} / \textbf{58.57}  &  94.62 / 97.55 \\
Degree & 69.31 / \textbf{56.88}  &  44.32 / \textbf{52.19}  &  54.53 / 26.64  &  70.23 / 70.4  &  2.36 / 12.31  &  25.48 / 54.2  &  97.12 / 98.79 \\
PageRank & 69.03 / 53.97  &  44.23 / 52.1  &  54.56 / 33.8  &  69.66 / \textbf{71.66}  &  2.88 / 11.94  &  45.41 / 50.86  &  96.98 / 98.79 \\
LG\{CoreNumber\} & 67.73 / 38.38  &  22.66 / 30.09  &  47.63 / 33.75  &  71.87 / 61.71  &  1.27 / 11.68  &  40.44 / 47.07  &  96.69 / 98.79 \\
LG\{Degree\} & 66.9 / 42.84  &  26.43 / 34.75  &  46.85 / 38.67  &  71.99 / 62.49  &  2.3 / 14.38  &  41.39 / 46.38  &  95.8 / \textbf{98.82} \\
LG\{PageRank\} & \textbf{70.87} / 37.46  &  24.82 / 35.13  &  49.5 / 37.83  &  \textbf{72.36} / 58.92  &  2.24 / 16.4  &  41.53 / 47.61  &  95.17 / 97.93 \\
\midrule
Default Ordering & 68.43 / 51.06  &  \textbf{52.78} / 48.48  &  \textbf{56.08} / 33.08  &  67.33 / 71.06  &  \textbf{4.43} / \textbf{30.72}  &  28.5 / 41.94  &  \textbf{98.25} / 95.51 \\

\bottomrule
\end{tabular}}
\caption{Accuracy scores for all tasks on the \textbf{GraphWave} (top) and \textbf{GraphQA} (bottom) datasets using \textbf{Llama 3 8B}. For each task, we compare the default labeling scheme, as provided by the corresponding graph generator, against the default order the edges have been generated. Notations remain the same as in Table \ref{table:graphwave_alltasks}.}
  \label{tab:defaultlabels}
\end{table*}

\noindent \textbf{LLM Family Variation with Extended Context.}
To further examine the influence of model architecture and extended context capacity, we evaluated the performance of Qwen 2.5 14B-1M ~\cite{qwen2.5-1m}, a large-context language model from a distinct model family capable of processing input sequences of up to 1 million tokens.
This evaluation allows us to assess whether architectural differences impact performance on graph-based reasoning tasks.
Table~\ref{table:qwen} reports the results obtained for the same selection of tasks previously used in Table~\ref{table:graphwave_70b}, enabling a direct comparison across models with varying capacities.

\begin{table}[ht]
  \resizebox{\columnwidth}{!}{
\begin{tabular}{@{}lcccc@{}}
\toprule
& \textbf{Node Counting}  &  \textbf{Node Degree}  &  \textbf{Diameter}  &  \textbf{Motifs' Shape} \\
\midrule
\textbf{Random Labels} &  &  &  & \\
CoreNumber & 70.67 / 72.12  & 76.5 / 68.66  &  12.1 / 12.5  &  58.5 / 56.3 \\
Degree & 71.23 / 70.96  &  76.97 / 66.66  &  12.07 / 13.3  &  \underline{62.27} / 59.43 \\
PageRank & 69.77 / 66.99  &  \underline{79.53} / \underline{72.09}  &  12.77 / 12.04  &  56.9 / 58.5 \\
LG\{CoreNumber\} & 71.13 / 68.92  &  72.17 / 54.08  &  12.63 / 14.17  &  59.17 / \textbf{\underline{61.97}} \\
LG\{Degree\} & 71.77 / 72.86  &  74.8 / 62.49  &  \underline{13.97} / \textbf{\underline{15.37}}  &  60.5 / 60.6 \\
LG\{PageRank\} & \underline{76.53} / \underline{76.59}  &  72.07 / 65.19  &  13.07 / 12.44  &  51.9 / 52.77 \\
\midrule
\textbf{Node Relabeling} &  &  &   & \\
CoreNumber &  \textbf{\underline{86.73}} / \textbf{\underline{87.66}}  & \textbf{\underline{80.23}} / \textbf{\underline{75.49}} &  12.93 / \underline{14.5}  &  58.03 / 53.17 \\
Degree & 74.2 / 77.63  & 78.03 / 72.59  &  12.8 / 11.3  &  61.3 / 49.83 \\
PageRank & 73.4 / 75.23 & 79.3 / 73.66 &  14.33 / 13.14  &  59.73 / 49.83 \\
LG\{CoreNumber\} & 76.03 / 71.52 &  78.77 / 69.19  &  12.63 / 12.64  & \textbf{\underline{64.03}} / \underline{57.53} \\
LG\{Degree\} & 80.03 / 81.39 &  76.6 / 69.22  &  \textbf{\underline{15.4}} / 12.77  &  60.57 / 46.77 \\
LG\{PageRank\} & 84.0 / 84.29 &  74.8 / 67.96 &  14.47 / 12.34  &  52.53 / 44.2 \\
\midrule
\midrule
\textbf{Baseline} & 78.84 / 77.43 & 70.09 / 69.19   &  9.95 / 10.74  &  54.53 / 57.3 \\
\bottomrule
\end{tabular}}
\caption{Accuracy scores for a subset of tasks on the \textbf{GraphWave} dataset using \textbf{Qwen 2.5 14B - 1M} as an ablation study. Notations remain the same as in Table \ref{table:graphwave_alltasks}.}
\label{table:qwen}
\end{table}

Although scaling from LLaMA 8B to 70B yields substantial gains in tasks such as node counting and motif shape classification, Table~\ref{table:qwen} demonstrates that Qwen 2.5 14B—despite having fewer parameters than LLaMA 70B—achieves competitive, and in some cases superior, performance across several tasks.
This is particularly evident in motif shape classification and diameter estimation, where Qwen's results rival or exceed those of the larger model.
Nonetheless, in line with trends observed across LLaMA variants, diameter estimation remains a consistently challenging task, with overall accuracy remaining low regardless of model architecture or scale.

\section{Performance and Complexity} \label{sec:complexity}

All considered graph measures, such as core number, degree, and PageRank, are computationally efficient.
For graphs where the number of edges exceeds the number of nodes, the computational complexity scales linearly with the number of edges, that is, \(\mathcal{O}(m)\), where \(m\) denotes the number of edges. Given that the graphs in the evaluated datasets are relatively small, the computation time required for these measures is negligible compared to the response generation time of the LLMs.

To illustrate this, we report on Table~\ref{table:complexity} the number of tokens generated per task along with the average time taken to produce a single response using LLama 3 8b.
Numerical tasks, LLM responses are concise and rapidly converge to a final answer.
In contrast, more complex tasks elicit longer responses that often involve intermediate reasoning steps.

\begin{table}[ht]
\centering
  \resizebox{\columnwidth}{!}{
\begin{tabular}{lcc}
\toprule
\textbf{Task} & \textbf{Number of Tokens} & \textbf{Inference Time (sec)} \\
\midrule
Node Counting   & 16  & 0.6 \\
Max Degree      & 16  & 0.6 \\
Node Degree     & 16  & 0.6 \\
Edge Existence  & 128 & 4.8 \\
Diameter        & 128 & 4.8 \\
Shortest Path   & 128 & 4.8 \\
Path Existence  & 128 & 4.8 \\
Motif’s Shape   & 16  & 0.6 \\
\bottomrule
\end{tabular}}
\caption{Tokens generated and response time for a single graph using LLaMA 3 8B across tasks.}
\label{table:complexity}
\end{table}

\section{Graph Size Limitations} \label{sec:graph_length}
In our approach, the entire graph is linearized into a token sequence and embedded directly into the model's input prompt.
As a result, the maximum size of the graph that can be processed in a single prompt is constrained by the model's context window.
Since the sequence length is primarily determined by the number of edges, we estimate the maximum number of edges that can be encoded per prompt for each model considered in this study.

To compute these estimates, we assume that each edge requires approximately 5 tokens to represent, and that the accompanying task description consumes an average of 100 tokens. Under these assumptions, Table~\ref{table:llm_max_edges} reports the estimated edge capacity corresponding to the context window of each model.

\begin{table}[ht]
  \resizebox{\columnwidth}{!}{
\begin{tabular}{@{}lccc@{}}
\toprule
\textbf{LLM} & \textbf{Context Length (tokens)} & \textbf{Max Number of Edges} \\
\midrule
\textit{LLama 3 8b} & 8,192 & 1,618\\
\textit{LLama 3 70b} & 8,192 & 1,618 \\
\textit{Qwen 2.5 14B 1M} & 1,010,000 & 199,980 \\
\bottomrule
\end{tabular}}
\caption{Estimated Maximum Number of Edges considering Context Window Size.}
\label{table:llm_max_edges}
\end{table}

In Table~\ref{table:dataset-stats}, we present statistics for several widely used graph datasets.
Many of these graphs are sufficiently small to fit within the context window of contemporary LLMs, with notable exceptions such as large-scale social and e-commerce networks (e.g., Reddit, Amazon).
This indicates that a substantial portion of benchmark graph datasets can be fully serialized and input to an LLM in a single prompt.
Nevertheless, in practical applications, even graph neural networks often rely on sampling strategies rather than processing entire graphs at once.
A similar strategy may be necessary when using LLMs for real-world graph tasks, depending on the application and scale.

Although our study primarily investigates the ability of LLMs to understand and reason over complete graph structures, we recognize that some of the tasks examined—such as node counting—are primarily diagnostic and may have limited practical relevance.
These tasks are intended to serve as controlled benchmarks to assess the reasoning capabilities of LLMs, rather than to reflect typical graph processing workloads.

\begin{table}[!ht]
\resizebox{\columnwidth}{!}{
\begin{tabular}{@{}lcc@{}}
  \toprule
  \textbf{Dataset (Source)}                                & \textbf{\#Nodes} & \textbf{\#Edges} \\ \midrule
  Cora~\cite{kipf2016semi}                             & 2,708           & 1,433            \\
  Citeseer~\cite{kipf2016semi}                         & 3,327           & 9,104            \\
  PPI~\cite{zitnik2017predicting}                           & 2,245           & 61,318           \\
  PubMed~\cite{kipf2016semi}                           & 19,717          & 88,648           \\
  Amazon~\cite{yang2018benchmarking}                               & 12,752          & 491,722          \\
  OGBN-Proteins~\cite{hu2020open}                          & 132,534         & 39,561,252       \\
  Reddit~\cite{hamilton2017inductive}                           & 232,965         & 114,615,892      \\
  Amazon Products~\cite{chiang2019cluster}                    & 1,569,960       & 264,339,468      \\
  \bottomrule
\end{tabular}
}
\caption{Node and edge counts for commonly used graph datasets.}
\label{table:dataset-stats}
\end{table}

\section{Prompt templates} \label{sec:prompts}

\noindent \textbf{Node Counting} \\
In an undirected graph $G$, $(i, j)$ means that node $i$ and node $j$ are connected with an undirected edge. \\
\textbf{Q:} How many nodes are in $G$?  \\
\textbf{G:} \{linearized graph\}

\medskip

\noindent \textbf{Max Degree} \\
In an undirected graph, $(i, j)$ means that node $i$ and node $j$ are connected with an undirected edge.  
The degree of a node is the number of edges connected to the node.  
Given a graph $G$ and its list of edges, respond to the following question:  \\
\textbf{Q:} Without any justification, what is the maximum node degree in the following graph $G$?  \\
\textbf{G:} \{linearized graph\}

\medskip

\noindent \textbf{Node Degree} \\
In an undirected graph, $(i, j)$ means that node $i$ and node $j$ are connected with an undirected edge.  
The degree of a node is the number of edges connected to the node.  
Given a graph $G$ and its list of edges, respond to the following question: \\
\textbf{Q:} Without any justification, what is the degree of node \{node\} in the following graph $G$?  \\
\textbf{G:} \{linearized graph\}

\medskip

\noindent \textbf{Edge Existence} \\
In an undirected graph, $(i, j)$ means that node $i$ and node $j$ are connected with an undirected edge.  
Given a graph $G$ and its list of edges, respond to the following question:  \\
\textbf{Q:} Does an undirected edge $(\{node1\}, \{node2\})$ exist in the following graph $G$?.  \\
\textbf{G:} \{linearized graph\} 

\medskip

\noindent \textbf{Diameter} \\
In an undirected graph, $(i, j)$ means that node $i$ and node $j$ are connected with an undirected edge.  
The diameter of a graph is the length of the shortest path between the most distanced nodes.  
Given a graph $G$ and its list of edges, respond to the following question:  \\
\textbf{Q:} Without any justification, what is the diameter of the following graph $G$?  \\
\textbf{G:} \{linearized graph\} 

\medskip

\noindent \textbf{Shortest Path} \\
In an undirected graph, $(i, j)$ means that node $i$ and node $j$ are connected with an undirected edge.  
Given a graph $G$ and its list of edges, respond to the following question:  \\
\textbf{Q:} Without any justification, what is the length of the shortest path from node $\{node1\}$ to node $\{node2\}$? If no path exists, the response is '0'.  \\
\textbf{G:} \{linearized graph\}  

\medskip

\noindent \textbf{Path Existence} \\
In an undirected graph, $(i, j)$ means that node $i$ and node $j$ are connected with an undirected edge.  
Given a graph $G$ and its list of edges, respond to the following question:  \\
\textbf{Q:} Does a path that connects node $\{node1\}$ and $\{node2\}$ exist in the following graph $G$?  \\
\textbf{G:} \{linearized graph\}

\medskip

\noindent\textbf{Motifs’ Shape Classification:} \\
In an undirected graph, $(i, j)$ means that node $i$ and node $j$ are connected with an undirected edge. The graph contains a motif graph with strictly one of the following structures. \{structure\}: \{definition\} \\
\textbf{Q: }Which of the defined structures is included in the following graph? \\
\textbf{graph:} \{linearized graph\}

\end{document}